\documentclass[conference]{IEEEtran}

\usepackage[utf8]{inputenc}
\usepackage{lmodern}
\usepackage{import}
\usepackage{cite}
\usepackage{amsmath,amssymb,amsfonts}
\usepackage{mathtools}
\usepackage{algorithm,algpseudocode}
\usepackage{graphicx}
\usepackage{textcomp}
\usepackage{xcolor}
\usepackage{pgfplots,pgfplotstable,subcaption}
\usepackage{booktabs}
\usepackage[usestackEOL]{stackengine}
\def\BibTeX{{\rm B\kern-.05em{\sc i\kern-.025em b}\kern-.08em
T\kern-.1667em\lower.7ex\hbox{E}\kern-.125emX}}

\algrenewcommand\algorithmicindent{1.0em}%

\begin{document}
\title{Component-Wise Natural Gradient Descent - An Efficient Neural Network Optimization}
\author{\IEEEauthorblockN{1\textsuperscript{st} Tran Van Sang}
\IEEEauthorblockA{\textit{The University of Tokyo}\\
Japan \\
0000-0002-4211-049X}
\and
\IEEEauthorblockN{2\textsuperscript{nd} Mhd Irvan}
\IEEEauthorblockA{\textit{The University of Tokyo}\\
Japan \\
irvan@yamagula.ic.i.u-tokyo.ac.jp}
\and
\IEEEauthorblockN{3\textsuperscript{rd} Rie Shigetomi Yamaguchi}
\IEEEauthorblockA{\textit{The University of Tokyo}\\
Japan \\
0000-0002-6359-2221}
\and
\IEEEauthorblockN{4\textsuperscript{th} Toshiyuki Nakata}
\IEEEauthorblockA{\textit{The University of Tokyo}\\
Japan \\
0000-0001-6383-7105}
}
\maketitle
\begin{abstract}
	Natural Gradient Descent (NGD) is a second-order neural network training that preconditions the gradient descent with the inverse of the Fisher Information Matrix (FIM).
	Although NGD provides an efficient preconditioner,
	it is not practicable due to the expensive computation required when inverting the FIM.
	This paper proposes a new NGD variant algorithm named Component-Wise Natural Gradient Descent (CW-NGD).
	CW-NGD is composed of 2 steps.
	Similar to several existing works,
	the first step is to consider the FIM matrix as a block-diagonal matrix whose diagonal blocks correspond to the FIM of each layer's weights.
	In the second step, unique to CW-NGD,
	we analyze the layer's structure and further decompose the layer's FIM into smaller segments whose derivatives are approximately independent.
	As a result, individual layers' FIMs are approximated in a block-diagonal form that trivially supports the inversion.
	The segment decomposition strategy is varied by layer structure.
	Specifically, we analyze the dense and convolutional layers and design their decomposition strategies appropriately.
	In an experiment of training a network containing these 2 types of layers,
	we empirically prove that CW-NGD requires fewer iterations to converge compared to the state-of-the-art first-order and second-order methods.
\end{abstract}

\section{Introduction}\label{sec:introduction}

Recently, Machine Learning has been rapidly growing and poses an inevitable role in modern society.
Among various machine learning architectures, Neural Network is the most powerful and versatile architecture that commits to the success of machine learning nowadays.
Especially since 2013, when ResNet~\cite{resnet-v2}, the first successful deep neural network, was introduced, Neural Network has been intensively investigated with regard to the number of layers, architectures, network and complexity,
becoming one of the best-performing architectures in Machine Learning.
Neural Network has been used in a wide range of applications, including image recognition, text recognition, speech recognition, natural language processing, machine translation, computer vision, and many others.
Along with the involvement of a new network model, a crucial component in neural networks that attracts much researcher's attention is the network training (optimization) algorithm,
the research field investigating algorithms that finds the optimal weights of the network.

Gradient Descent (GD) is a popular network optimization family that optimizes the weight by following the steepest descent direction of the loss function.
First-order GD methods precondition the direction with a scalar learning rate, usually defined via an adaptive strategy~\cite{adam,rmsprop,adadelta,adagrad,nesterov,nadam},
while second-order GD methods precondition the direction with the inverse of the Hessian matrix.
Natural Gradient Descent (NGD)~\cite{ngd} is a second-order GD variant that uses the Fisher Information Matrix (FIM)~\cite{fim-def} in the place of the Hessian.
A great number of works~\cite{ngd-perf-1,ngd-perf-2,ngd-perf-3,osawa,kfac-1,kfac-2,kfac-3,kfac-4} have shown that NGD and its variants produce faster convergence than first-order GD.
Besides, NGD requires fewer hyperparameters, thus simplifying the tuning process.
However, the application of NGD is limited due to the burdensome calculation necessary for inverting the FIM.
Reducing the amount of calculation in NGD and making it applicable in practice is an open active area of research.
Tackling the challenge, we investigate a novel method named Component-Wise Natural Gradient Descent (CW-NGD),
an improved variant of NGD that produces high accuracy results and requires a reasonable running time.

CW-NGD is composed of 2 steps.
The first step, similar to several existing works~\cite{kfac-1,kfac-2,kfac-3,kfac-4,osawa,block-wise-ngd},
is to consider the FIM matrix as a block-diagonal matrix whose diagonal blocks correspond to the FIM of each layer's weights.
In the second step, which is unique to CW-NGD,
for dense and convolutional layers,
we analyze the layer's structure and further decompose the layer's FIM into smaller segments whose derivatives are approximately independent.
As a result, individual layers' FIMs are approximated in a block-diagonal form that trivially supports the inversion.

In an experiment with a network containing convolutional and dense layers on the MNIST dataset,
CW-NGD converges within fewer iterations than the best-known first-order and second-order optimization algorithms, Adam~\cite{adam} and KFAC~\cite{kfac-1,kfac-2,kfac-3,kfac-4}, respectively.

In the next section, Section~\ref{sec:related-works}, we discuss several existing works in the literature before describing our method in Section~\ref{sec:methodology}.
The experiment detail is provided in Section~\ref{sec:experiment}.
Next, we discuss the results of our method in Section~\ref{sec:results}.
Finally, we summarize and discuss the future of our method in Section~\ref{sec:conclusion}.

\section{Related Works}\label{sec:related-works}

First-order optimization involves the use of a learning rate adaptive strategy:
Adam, AdaMax, RMSProps, AdaDelta, AdaGrad, SGD with Nesterov's accelerated gradient, Nadam~\cite{adam,rmsprop,adadelta,adagrad,nesterov,nadam}, etc.
Some of them, for instance, Adam, AdaGrad~\cite{adam}, use the moving average of squared gradients as a heuristic of the FIM,
the matrix contains the quadratic curvature of the loss function.
Their prominent usage in practice and well-performed results highly promote the usefulness of the use of quadratic curve information in network optimization.

The initial idea of optimizing networks using quadratic curvature and dividing the network parameters into derivative-independent components was first proposed by Kurita et al.~ in 1993~\cite{kurita-1993}.
The motivation of the authors was to apply the iterative weighted least squares algorithm,
which is usually used in Generalized Linear Model training,
to network optimization.
However, this approach supports only dense networks with a single hidden layer, and the experiment was on a simple XOR problem that has not been commonly used recently.
Nonetheless, it inspired the recently introduced unit-wise FIM approximation~\cite{ollivier-2015}.

NGD~\cite{ngd}, proposed in 1998, later became the most prominent variant of the second-order method in Neural Network optimization.
NGD uses the FIM as the preconditioner of the gradient descent update in the place of the Hessian matrix.
While the Hessian preconditioner optimizes the Euclidean distance,
the FIM preconditioner in NGD is created to adapt to the information geometry.
It aims to minimize the distance in the probabilistic space, whose distance is defined by the Kullback-Leibler divergence.
Although the initial motivation of NGD is not to approximate the Hessian,
it was later proven that the FIM coincides with a generalized Gauss-Newton~\cite{ggn} approximation of the Hessian under overwhelmingly practical cases~\cite{fisher-limit}.
This coincidence explicitly implies the classification of NGD as a second-order optimization method.

NGD can produce a more efficient learning step compared to almost any first-order optimization method.
However, the FIM inverting operation in NGD requires a significant computation time, which restricts the application of NGD to simple neural network models.
To remedy this problem, several existing works~\cite{tonga,ollivier-2015,amari-2019,kfac-1,kfac-2,kfac-3,kfac-4,block-wise-ngd} have proposed various approximation schemes of the FIM for a cheap inversion.
For instance, in 2015, inspired by~\cite{kurita-1993} and theoretically evaluated by Amari et al.~\cite{amari-2019}, Ollivier et al.~\cite{ollivier-2015} extended NGD with the unit-wise NGD method.
The authors introduce a theoretical framework to build invariant algorithms,
in which they treat NGD as an invariant method that optimizes the distance in Riemannian geometry.
Based on the suggested framework,
the authors grouped the update by the output nodes and calculated individual group's FIM for each update, thus,
ignoring the interaction among the node groups.
However, the approach does not apply to the convolutional layer and is limited to simple neural network models due to the inefficient gradient calculation.

The more modern and practicable extension of NGD is KFAC, initially introduced in 2015 by James et al.~\cite{kfac-1,kfac-2,kfac-3,kfac-4}.
This is considered the state-of-the-art second-order method in neural network optimization.
KFAC works by approximating the off-diagonal blocks of the FIM as zero,
then factorizing the diagonal blocks as a Kronecker product of two smaller matrices.
The second step reduces the computational cost of inverting the FIM.
In contrast, it greatly reduces the accuracy of the approximation.

Besides NGD, which uses the FIM as the preconditioner,
another research direction of second-order optimization is to approximate the Hessian preconditioner.
The most well-known methods are Broyden Fletcher Goldfarb Shanno (BFGS) and L-BFGS methods~\cite{lbfgs}.
These methods work on any differentiable function and, in reality, are used to optimize the neural network loss function.
Instead of calculating the Hessian in every step,
the former works by updating the Hessian matrix calculated in previous iterations.
Although it is more computationally efficient than directly deriving the second-order derivative of the loss function,
it requires storing the update vector in each step, hence is very memory-consuming.
The latter was introduced to remedy that problem by only storing the update vector of k last iterations.
Nevertheless, these methods are not used in practice for large-scale problems (large networks or large datasets),
because they do not perform well in mini-batch settings and produce low-accuracy results~\cite{lbfgs-limit}.

In the case of dense layers,
CW-NGD is identical to unit-wise NGD introduced in~\cite{ollivier-2015,osawa,amari-2019}.
We explicitly declare our additional contributions as follows.
Firstly, CW-NGD analyzes the layer's structure to provide an efficient FIM-approximation scheme,
while unit-wise NGD works by grouping the network weights into groups by output nodes.
They accidentally turn out to be the same approximation scheme in the case of dense layers.
Secondly, CW-NGD is applicable to convolutional layers while unit-wise is not.
Thirdly, studies on unit-wise NGD~\cite{ollivier-2015,osawa,amari-2019} focus on theoretical analysis of the method and lack practical application analysis.
In contrast, CW-NGD is verified with the MNIST dataset~\cite{mnist},
which demonstrates the more potential practicability of CW-NGD over unit-wise NGD.

\section{Methodology}\label{sec:methodology}

In this section,
we first introduce several basic notations and the neural network definition in the first subsection.
In the next subsection, we briefly introduce the NGD method, FIM definition, and layer-wise NGD.
After that, we describe our proposed method Component-Wise Natural Gradient Descent (CW-NGD) in the following two subsections for the cases of dense and convolutional layers, respectively.
Finally, we provide an efficient implementation for CW-NGD, evaluate the computational performance gained by CW-NGD, and discuss the potential parallelization capability of the method.

To be consistent with the notation of derivatives with respect to vector and matrix, we consider a vector as a row of elements.
In other words, a vector from $\mathbb{R}^n$ space is equivalent to a $1\times n$ matrix in $\mathbb{R}^{1\times n}$.
\subsection{Neural Network Definition And Notations}\label{subsec:neural-network-definition-and-notations}

We consider a neural network consisting of $L$ dense layers and a predictive exponential family distribution $r(y|z)$.
Note: we refer to a distribution by its density function for convenience.
$r(y|z)$ family covers the most popular use cases, for instance, $r(y|z) = \mathcal{N}(y,z,\sigma^2)$ for least-squares regression, $r(y|z) = \sum_{c=1}^{c=C}y_c\frac{e^{z_c}}{\sum_{c=1}^{c=C}e^{z_c}}$ for C-class classification with $y$ is one-hot vector and $z_c$ is the $c$-th element of $z$.
The network transforms an input $x=a_0 \in \mathbb{R}^{n_0}$ to output $z=a_L \in \mathbb{R}^{n_L}$ through $L$ layers,
and obtain the $y$ output's probability by $r(y|z)$.
For $l = 1, 2, \ldots, L$, $s_l = a_{l-1} W_l$, $a_l = \phi_l (s_l)$
where $\phi_l (s_l)$ is a (non-linear) activation function,
$s_l \in \mathbb{R}^{n_l}$ is the pre-activation,
$a_l \in \mathbb{R}^{n_l}$ is the (post-)activation,
$W_l \in \mathbb{R}^{n_{l-1} \times n_{l}}$ is the weight matrix
(\textit{Note}: the bias term's effect can be achieved by
prepending a homogeneous value 1 to the activation vectors $a_l$).

The network parameter $\theta \in \mathbb{R}^{1 \times (n_0 n_1 + n_1 n_2 + \cdots + n_{L-1} n_L)}$ is defined by chaining all weight matrices $\theta=\left[\theta_1\cdots \theta_L\right]$
with $\theta_l=vec(W_l)$ is a vectorization of $W_l^\intercal$ composed by chaining rows of matrix $W_l$ into a single row-vector.
We define $\mathcal{D}_\theta = \nabla_\theta\left(-\log r_\theta(y|x)\right)$,
function $f(x,\theta)=z$,
and conditional distribution $r_\theta(y|x)=r(y|f(x,\theta))$ of the joint distribution $r_\theta(x,y)=r(x)r_\theta(y|x)$.
Given a dataset of $P$ pairs of samples and associated labels $(x_p,y_p)$,
let $\mathcal{D}_\theta^{(p)} = \nabla_\theta\left(-\log r_\theta(y_p|x_p)\right)$.

To easily identify elements, we use the notation $\theta_{(l,i,j)}$ (or $\mathcal{D}_{\theta,(l,i,j)}$) to refer to the element of the flattened vector $\theta$ (or $\mathcal D_\theta$)
corresponding to the cell of the $l$-th component matrix of $W_l$ (or $\mathcal D_{W_l}$) at $i$-th row and $j$-th column.
Similarly, for matrix $M$ (e.g., the FIM $\mathcal{F}_\theta$ introduced in the next subsection) whose both dimensions are equal to the vectorized vector's dimension,
we use the notation $M_{(l_1, i_1, j_1),(l_2,i_2,j_2)}$ (e.g., $\mathcal{F}_{\theta,(l_1,i_1,j_1),(l_2,i_2,j_2)}$) to refer to the cell of $M$ at $(l_1,i_1,j_1)$-th row and $(l_2,i_2,j_2)$-th column.
\subsection{Fisher Information Matrix And Natural Gradient Descent}\label{subsec:fisher-information-matrix-and-natural-gradient-descent}
This subsection begins with the definitions of NGD and FIM.
Then it describes the method of approximating FIM via a layer-wise block-diagonal matrix.

Gradient Descent is a family of network optimization methods that optimizes the weight by following the steepest descent, equivalently, the negative gradient, direction of the loss function.
First-order optimizations precondition the direction with a scalar learning rate.
Second-order optimizations precondition the direction with the inverse of the Hessian matrix.
NGD\cite{ngd} is a second-order optimization variant that uses the FIM\cite{fim-def} $\mathcal{F}_\theta$'s inverse as the preconditioner.
\begin{align}
	\mathcal F_\theta &= \textstyle \sum_{p=1}^{p=P}\left({\mathcal D_\theta^{(p)}}^\intercal \mathcal D_\theta^{(p)}\right)\label{eq:fim} \\
	\mathcal F_\theta^{(stat)} &= P \mathbb E_{x,y\sim r(x,y)}\left[\mathcal D_\theta^\intercal \mathcal D_\theta\right] \label{eq:fim-stat} \\
	\mathcal F_\theta^{(EF)} &= \textstyle \sum_{p=1}^{p=P} \mathbb E_{y\sim r(y|x_p)} \left[\mathcal D_\theta|_{x=x_p}^\intercal \mathcal D_\theta|_{x=x_p}\right]\label{eq:fim-ef}
\end{align}
There is usually confusion among the 3 definitions of the FIM~\cite{fisher-limit}.
The definition given by Eq.~\ref{eq:fim-ef} is called Empirical FIM (EF) by statisticians~\cite{fim-stat-1,fim-stat-2}
but is called FIM in Machine Learning literature~\cite{fim-ml-1,fim-ml-2}.
On the other hand, Eq.~\ref{eq:fim-stat} is what statisticians call the FIM,
while Eq.~\ref{eq:fim} is called EF in Machine Learning.
We explicitly use the definition in Eq.~\ref{eq:fim} as FIM in this work for simplicity.

NGD provides an efficient preconditioner for the gradient descent in practice~\cite{ngd-perf-1,ngd-perf-2,ngd-perf-3,osawa,kfac-1,kfac-2,kfac-3,kfac-4}.
However, its usage is limited due to the expensive calculation required to invert the FIM.
The common workaround is to approximate the FIM as a block-diagonal matrix whose diagonal blocks are the FIM of each layer $\mathcal F_{\theta,l}$, such as in KFAC~\cite{kfac-1,kfac-2,kfac-3,kfac-4}.
\begin{align*}
	\mathcal F_{\theta} = diag\left(\mathcal F_{\theta,1}, \cdots, \mathcal F_{\theta,L}\right) \\
	\mathcal F_{\theta}^{-1} = diag\left(\mathcal F_{\theta,1}^{-1}, \cdots, \mathcal F_{\theta,L}^{-1}\right)
\end{align*}
with $diag(A_1,A_2,\cdots,A_n)$ is a block diagonal matrix whose diagonal blocks are $A_1, A_2, \ldots, A_n$ in the given order.
By this approximation,
instead of inverting one large $\mathcal F_\theta$,
we can invert individual layer FIMs $\mathcal F_{\theta,l}$ ($l=1,2,\ldots,L$) and aggregate them into a block-diagonal matrix.
However, in practice, $\mathcal F_{\theta,l}$ is usually large and still expensive to invert.
KFAC~\cite{kfac-1,kfac-2,kfac-3,kfac-4} remedies this by decomposing the layer FIM as a Kronecker product of smaller vectors,
but it produces a low-accuracy approximation.

In the next subsections, we propose our method named Component-Wise Natural Gradient Descent (CW-NGD),
a novel precise and practicable approximation of the layer FIMs which also supports the inversion trivially.
Specifically, subsections~\ref{subsec:approximate-fim-inversion-for-dense-layer} and \ref{subsec:approximating-the-fim-for-convolutional-layers} describe the method that applies to dense layers and convolutional layers, respectively.
\begin{figure*}
	\centering
	\vspace{-1cm}
	\begin{subfigure}{\textwidth}
		\includegraphics[width=\textwidth,trim={0 19.7cm 0 0},clip]{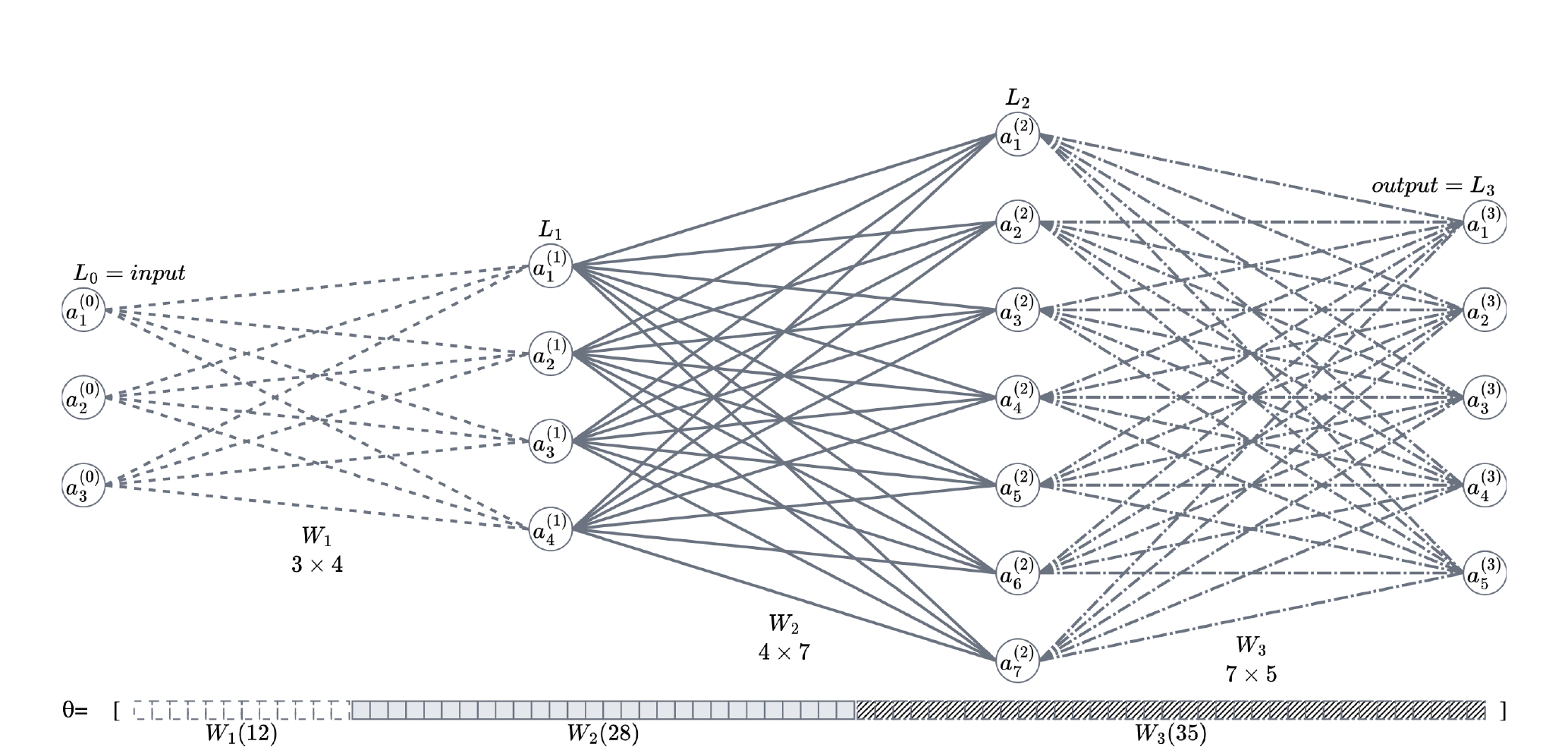}
		\caption{Sample dense network}\label{fig:sample-dense-net}
	\end{subfigure}

	\begin{subfigure}{.45\textwidth}
		\includegraphics[width=\textwidth,trim={0 11cm 0 0},clip]{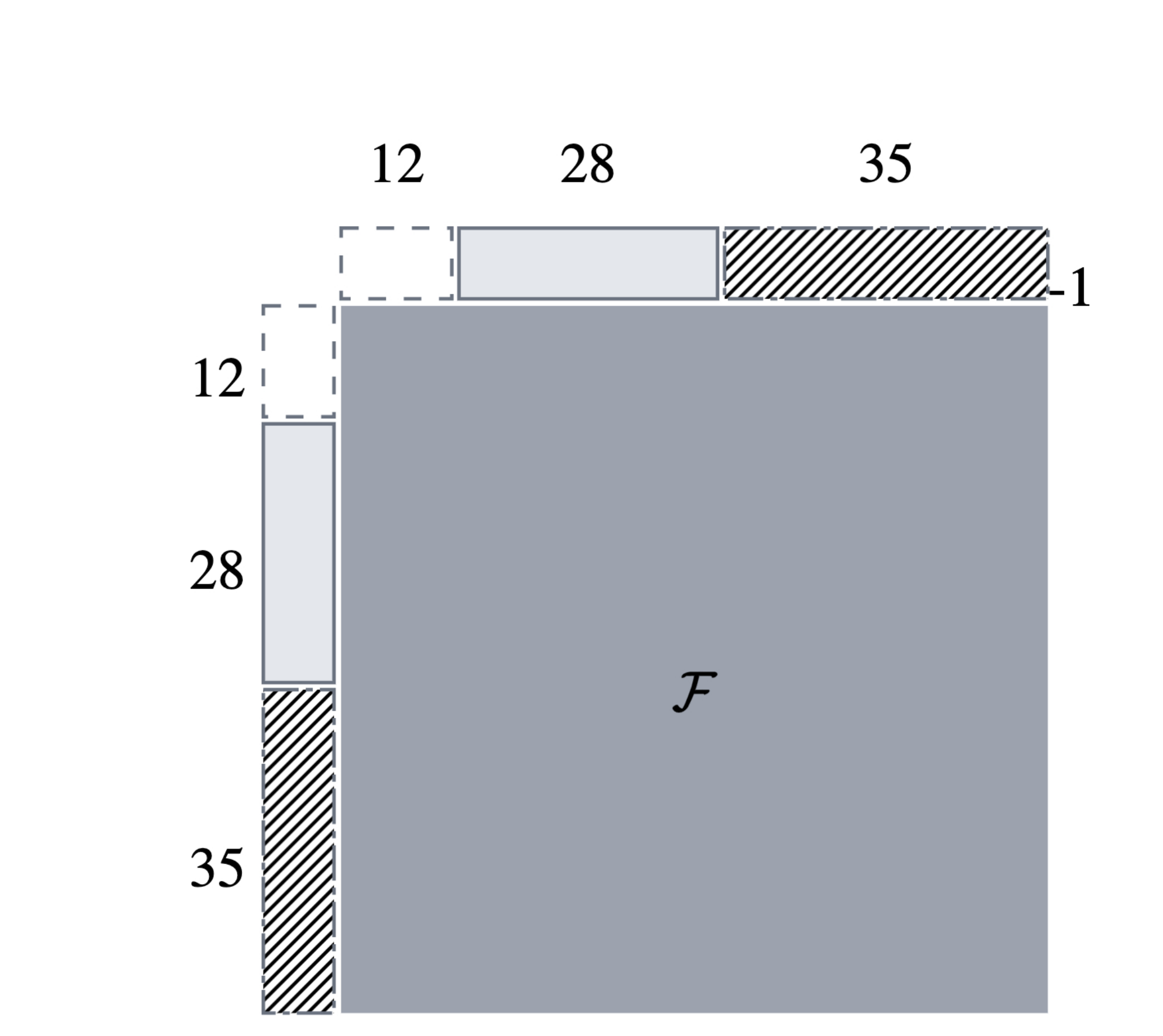}
		\caption{Full FIM}\label{fig:full-fim}
	\end{subfigure}
	\begin{subfigure}{.45\textwidth}
		\includegraphics[width=\textwidth,trim={0 11cm 0 0},clip]{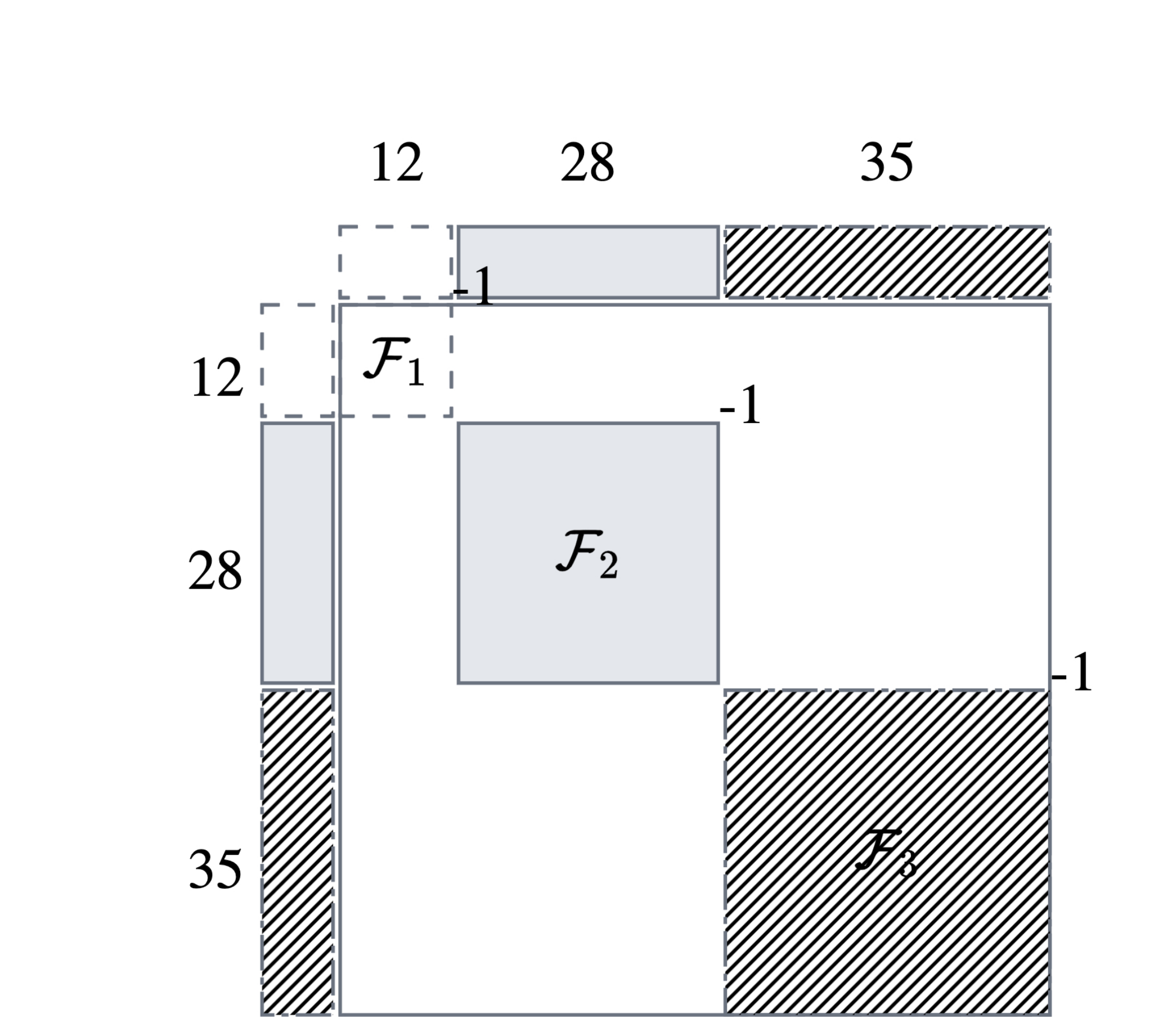}
		\caption{Layer-Wise FIM}\label{fig:layer-fim}
	\end{subfigure}

	\begin{subfigure}{.45\textwidth}
		\includegraphics[width=\textwidth,trim={0 11cm 0 0},clip]{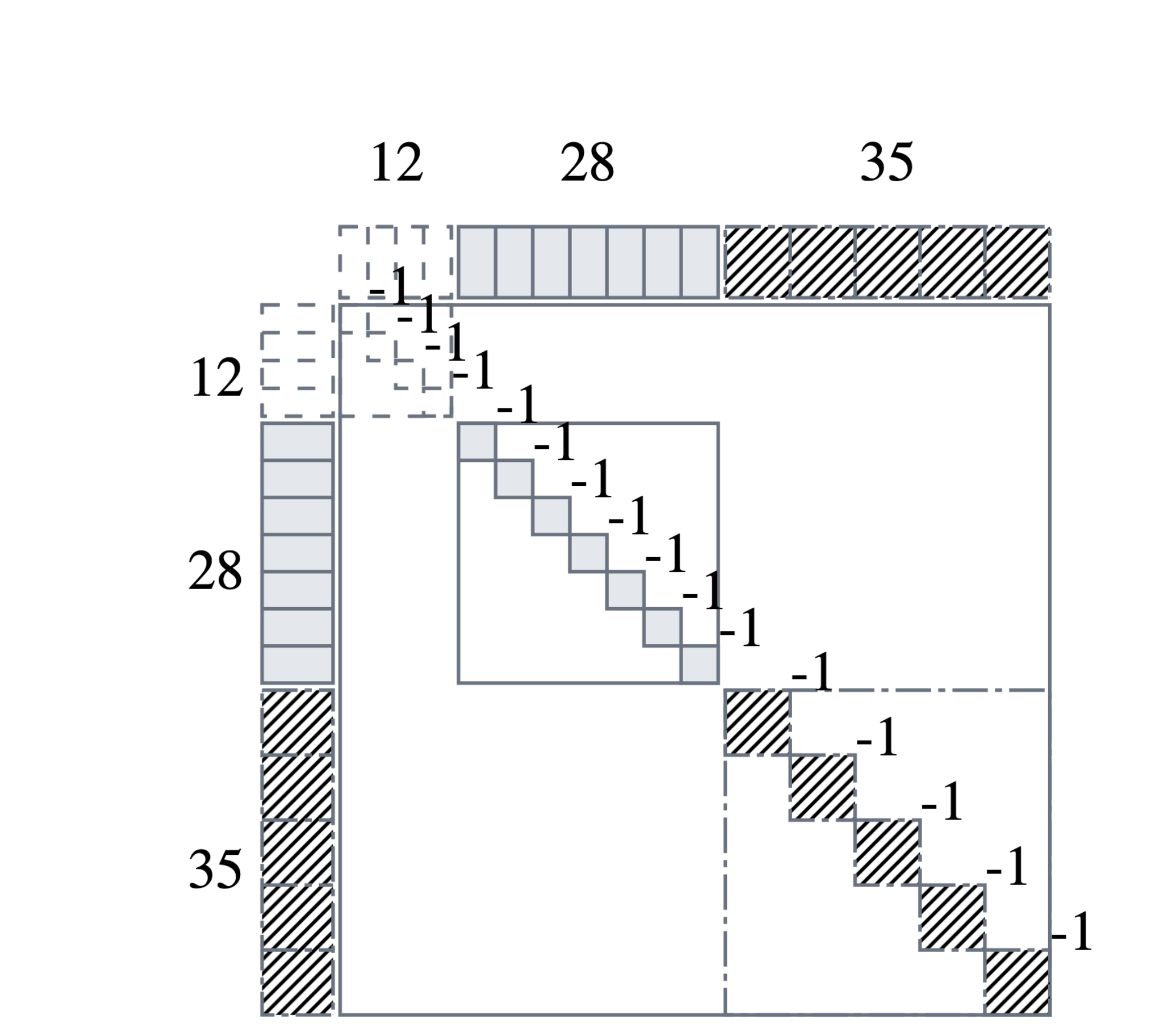}
		\caption{CW-NGD: Component-Wise FIM}\label{fig:component-fim}
	\end{subfigure}
	\begin{subfigure}{.45\textwidth}
		\includegraphics[width=\textwidth,trim={0 11cm 0 0},clip]{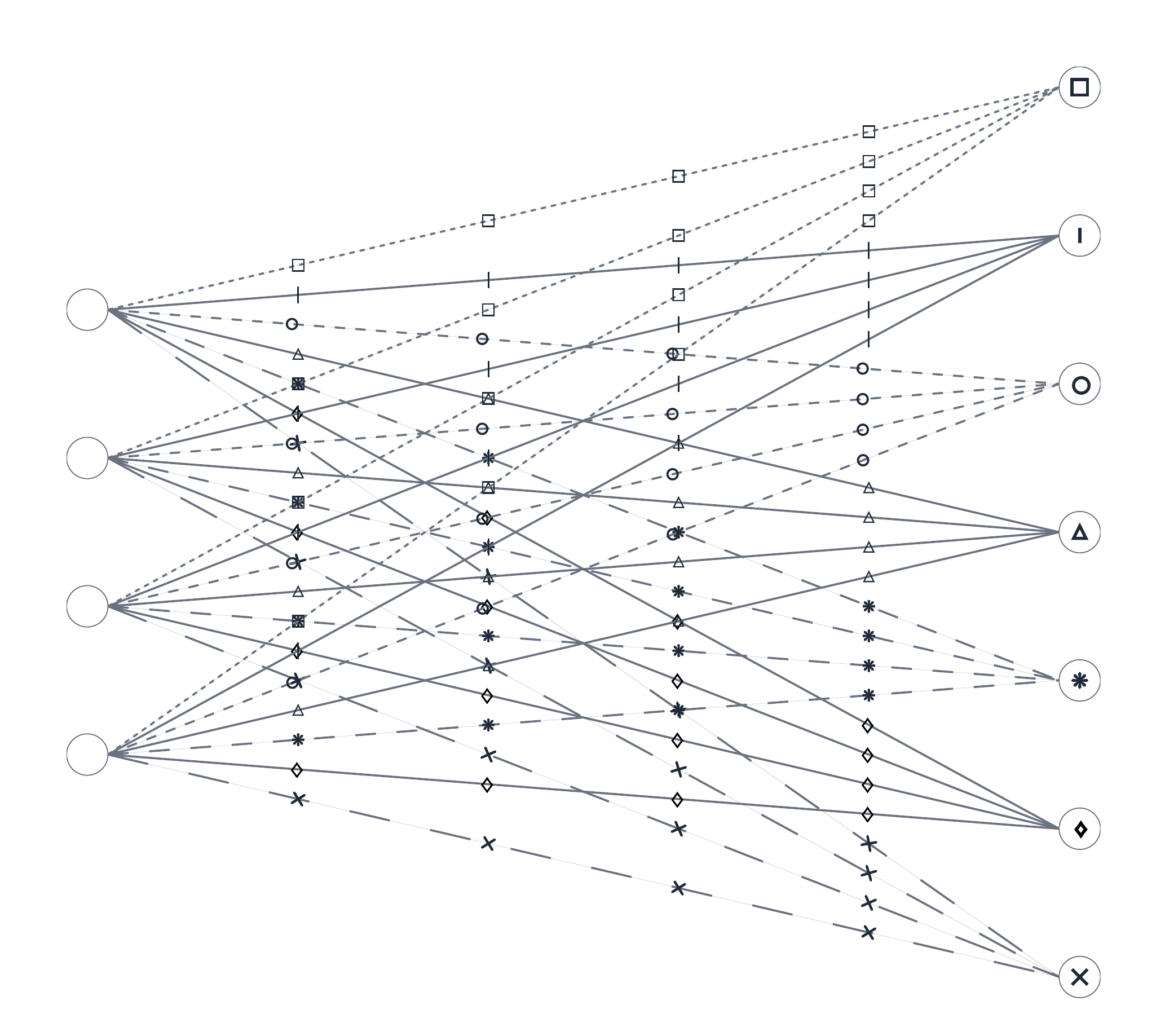}
		\caption{Independent components}\label{fig:independent-component}
	\end{subfigure}
	\caption{The approximation of a dense layer's FIM and its inverse}
\end{figure*}

\subsection{Approximate FIM Inverse for Dense Layer}\label{subsec:approximate-fim-inversion-for-dense-layer}

For the sake of  transparent interpretation,
we consider a simple dense network illustrated in Figure~\ref{fig:sample-dense-net}.
The original NGD algorithm requires inverting the FIM depicted in Figure~\ref{fig:full-fim}.
Figure~\ref{fig:layer-fim} represents the Layer-Wise NGD algorithm family's mechanism,
where they approximate the FIM as a block-diagonal matrix and inverts each diagonal block.
In CW-NGD, we divide the weight vector $\theta_l$ into groups in which elements of a group share the same output node (Figure~\ref{fig:independent-component}).
Namely, we divide $\theta_l$ (or $\mathcal{D}_{\theta,l}$) into $n_l$ groups of elements $\theta_{l,:,i}$ (or $\mathcal{D}_{\theta, (l,:,i)}$) ($i=1,2,\ldots,n_l$).
Thanks to the definition of the vectorization function $vec$ of the weight matrix,
elements in the same group are consecutive in the vectorized presentation of the weight.
This results in $\mathcal F_{\theta,l}$ being a $n_l$ by $n_l$ block matrix in which the $i$-th row and $j$-th block is $\mathcal{F}_{\theta,(l,:,i),(l,:,j)}$.
By definition of $\mathcal F_\theta$ given in Eq.~\ref{eq:fim}, we have:
\begin{align}
	\mathcal{F}_{\theta,(l,:,i),(l,:,j)} &= \sum_{p=1}^{p=P}\left({\mathcal D_{\theta,(l,:,i)}^{(p)}}^\intercal \mathcal D_{\theta,(l,:,j)}^{(p)}\right)\label{eq:fim-layer}
\end{align}

There is a well-known assumption that holds very well in practice~\cite{grad-indep-5} called gradient independence assumption in the mean field theory.
It is inspired by~\cite{grad-indep-inspire}, first introduced in~\cite{grad-indep-origin}, and later applied in numerous studies in Deep Neural Networks~\cite{grad-indep-1,grad-indep-2,grad-indep-3,grad-indep-5,grad-indep-6,grad-indep-7,osawa}.
Under this assumption, the network weight for backpropagation is assumed to be different from those used in forwarding propagation and identically independently sampled from the same distributions.
That means their covariance is zero.
For $i\neq j$, we have:
\begin{align}
	0 &= cov(\mathcal{D}_{\theta,(l,:,i)}, \mathcal{D}_{\theta,(l,:,j)}) \\
	0&= \mathbb{E}\left[ \mathcal{D}_{\theta,(l,:,i)}^\intercal \mathcal{D}_{\theta,(l,:,j)} \right] - \mathbb{E}\left[\mathcal{D}_{\theta,(l,:,i)}\right]^\intercal  \mathbb{E}\left[\mathcal{D}_{\theta,(l,:,j)}\right] \\
	0&= \mathbb{E}\left[ \mathcal{D}_{\theta,(l,:,i)}^\intercal \mathcal{D}_{\theta,(l,:,j)} \right] \label{eq:indep-derivative}
\end{align}
The last equality is based on the zero expectation property of the score function~\cite{score-function} ($\mathbb{E}\left[\mathcal{D}_{\theta,(l,:,i)}\right] = 0$) that satisfies under overwhelmingly the majority of the cases within the application of neural networks~\cite{rothman-regularity} and can be trivially proven using the Leibniz Integral Rule~\cite{leibniz}.
Evaluating the Eq.~\ref{eq:indep-derivative} with samples from dataset, we have $\sum_{p=1}^{p=P}\left({\mathcal{D}_{\theta,(l,:,i)}^{(p)}}^\intercal \mathcal{D}_{\theta,(l,:,j)}^{(p)} \right)= 0$ for $i\neq j$.
Correspondingly, off-diagonal blocks of layer FIM (Eq.~\ref{eq:fim-layer}) are zero.
As a result, the dense layer's FIM becomes a block-diagonal matrix whose inverse can be calculated by inverting its diagonal block sub-matrices within a reasonable time.
Figure~\ref{fig:component-fim} reveals the mechanism of CW-NGD.
\begin{align*}
	\mathcal{F}_{\theta,l} = diag(\mathcal{F}_{\theta,(l,:,1),(l,:,1)},\ldots, \mathcal{F}_{\theta,(l,:,n_l),(l,:,n_l)}) \\
	\mathcal{F}_{\theta,l}^{-1} = diag(\mathcal{F}_{\theta,(l,:,1),(l,:,1)}^{-1},\ldots, \mathcal{F}_{\theta,(l,:,n_l),(l,:,n_l)}^{-1})
\end{align*}
In summary, in the CW-NGD method, we apply the derivative independence assumption to pairs of edges associated with different output nodes (Figure~\ref{fig:independent-component}).
This division is based on the observation that elements across different groups do not directly contribute to the loss function at the $l$-th layer stage of the calculation in $f(x,\theta)$.
To put it in another way, 2 elements belonging to 2 different groups contribute to different output nodes whose values are passed through the activation function before contributing to the same output node in the next layer.
By the chaining rule which is used in backpropagation, the derivative with respect to the next layer's output node must be multiplied with the derivative with respect to different output nodes in the current layer before being used to calculate the derivative with respect to these 2 elements.
These analyses are further supported by our empirical result, which will be shown in Section~\ref{sec:results}.
\subsection{Approximating the FIM for Convolutional Layers}\label{subsec:approximating-the-fim-for-convolutional-layers}
\begin{figure*}
	\centering
	\begin{subfigure}{\textwidth}
		\includegraphics[width=\textwidth,trim={0 10.8cm 0 0},clip]{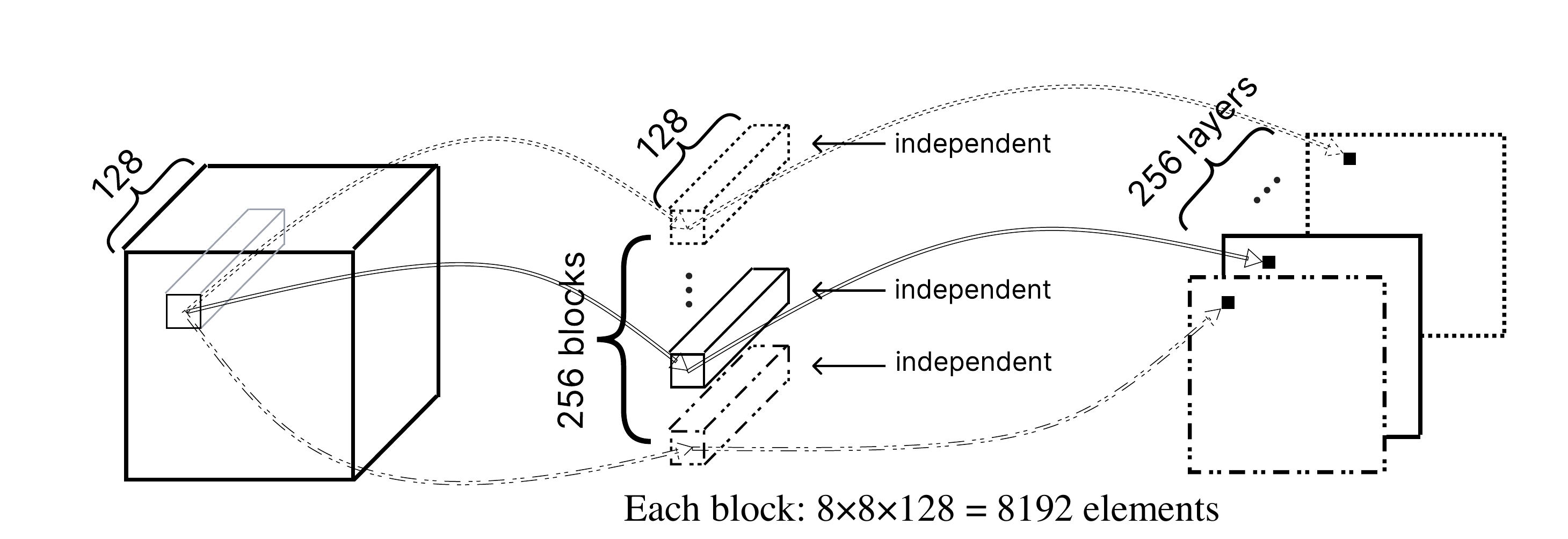}
	\end{subfigure}
	\caption{Sample CNN}\label{fig:sample-cnn}
\end{figure*}
\begin{figure}
	\centering
	\begin{subfigure}{0.4\textwidth}
		\includegraphics[width=\textwidth,trim={0 10.8cm 0 0},clip]{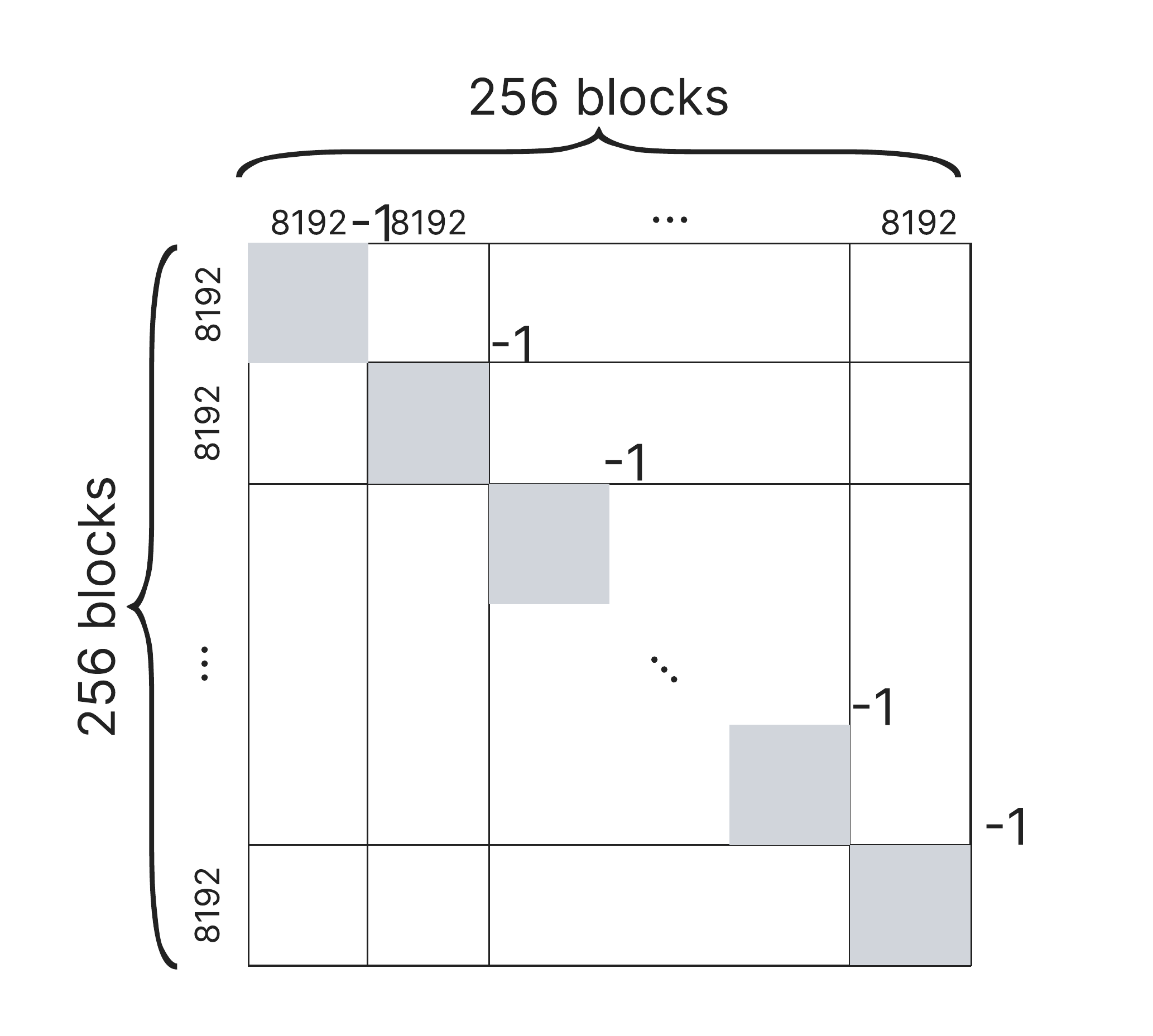}
	\end{subfigure}
	\caption{CW-NGD's FIM's approximation for CNN layers}\label{fig:cnn-fim}
\end{figure}
This subsection focuses on approximating the FIM for convolutional layers.
Figure~\ref{fig:sample-cnn} exposes the structure of a sample convolutional with filter size $8 \times 8$,
input layer having 128 channels,
output layer having 256 channels.
The array of blocks in the middle of Figure~\ref{fig:sample-cnn} represents the weight of the layer consisting of 256 blocks with each block having $8\times 8 \times 128 = 8192$ scalar elements.
Because convolutional layers have a different structure from dense layers,
we need to adapt the notation accordingly.
Most of the formulae are the same as of the dense layer case, though.

Suppose that between the $(l-1)$-th layer with $c_{l-1}$ channels and the $l$-th layer with $c_l$ channels,
there is a 2D convolutional layer with a $k_l\times e_l$-size kernel.
The convolutional layer has the weight tensor in a shape $k_l\times e_l \times c_{l-1} \times c_l$ and the bias in a shape $c_l$.
Diverging from the case of dense layer,
for the convolutional layer,
we use the index notation $\theta_{l,r,h,i,o}$ (or $\mathcal{D}_{\theta,(l,r,h,i,o)}$) to refer to the element of the flattened vector $\theta$ (or $\mathcal{D}_\theta$) associated with the element of the weight of the $l$-th (convolutional) layer at $(r,h,i,o)$ index with $r, h$ being kernel's indices and $i,o$ being input layer and output filter's indices, respectively.
For the bias element, we use the notations $\theta_{l,o}$ and $\mathcal{D}_{\theta, (l,o)}$ instead.

The associated layer FIM for this convolutional layer turns in a shape $(c_l+k_le_lc_{l-1}c_l)\times (c_l+k_le_lc_{l-1}c_l)$.
We index this FIM using the notations $\mathcal{F}_{\theta,(l_1,r_1,h_1,i_1,o_1),(l_2,r_2,h_2,i_2,o_2)}$, $\mathcal{F}_{\theta,(l_1,r_1,h_1,i_1,o_1),(l_2,o_2)}$, $\mathcal{F}_{\theta,(l_1,o_1),(l_2,r_2,h_2,i_2,o_2)}$, and $\mathcal{F}_{\theta,(l_1,o_1),(l_2,o_2)}$ for respective pairs of element types described using the index notation for the weight tensor $\theta$.

The convolutional layer structure does not affect the generality of the derivative independence assumption and its derivation to the block-diagonal matrix approximation of the FIM described in Subsection~\ref{subsec:approximate-fim-inversion-for-dense-layer}.
All we need to do is to provide an appropriate grouping scheme for an efficient approximation.
For the case of the dense layer, we divide the weight into groups by output node.
In contrast, for the convolutional layer, we divide the weight and bias into groups by their output layers.
In other words, $\mathcal D_{\theta,l}$ is divided into $c_l$ groups.
For $o=1,2\cdots,c_l$, the $o$-th group contains 1 element of $\mathcal D_{\theta,(l,o)}$ and $k_l e_l c_{l-1}$ elements from $\mathcal D_{\theta,(l,:,:,:,o)}$.
In the flattened representation of the weight tensor,
we place the elements in each group consecutively, beginning with the bias element.
Applying the derivative independence assumption,
the FIM becomes a block-diagonal matrix consisting of $c_l$ blocks, each of which is a $(1 + k_l e_l c_{l-1})\times (1 + k_l e_l c_{l-1})$ matrix.
In this form, the FIM can be inverted in a reasonable time.
Figure~\ref{fig:cnn-fim} shows how the FIM and its inverse are approximated for the case of the sample convolutional layer.

The division strategy for the convolutional layer is designed in the same way as the dense layer case.
In which elements belonging to different groups do not directly contribute to the forward propagation of the network at the same stage of the current layer.
To put it another way, their values contribute to different output nodes whose values are passed through the activation function before contributing to the same output node in the next layer.
Hence, in backpropagation,  the derivative with respect to the next layer's output node must be multiplied with the derivative of the activation function with respect to different output nodes in the current layer before being used to calculate the network loss's derivative with respect to these 2 elements.

The mechanism of CW-NGD for convolutional layers clearly clarifies our contribution compared to the unit-wise NGD method~\cite{ollivier-2015,amari-2019}.
Unit-wise NGD coincides with CW-NGD in the case of dense layers.
However, for convolutional layers,
the layer has $W_lH_lc_l$ output nodes and many of them share the same weights,
making unit-wise NGD impossible to group the weights by output nodes for update.

On the other hand,
regarding the derivative independence assumption,
it is theoretically possible to apply it to all pairs of different weight elements and yield FIM in form of a scalar diagonal matrix.
\cite{osawa} calls this method as entry-wise NGD, evaluates it on a crafted dataset and shows that it converges faster than the pure first-order methods (stochastic gradient descent), but slower than layer-wise NGD and unit-wise NGD.
In our proposed method CW-NGD, we apply the derivative independence assumption to selected pairs of elements based on the grouping strategy which is designed based on the layer structure.
Additionally, we empirically show that CW-NGD produces an efficient training convergence in a more practical dataset in Section~\ref{sec:results}.
\subsection{Efficient CW-NGD Implementation and Discussion}
\begin{algorithm}
	\caption{Component-Wise Natural Gradient Descent}\label{alg:cw-ngd}
	\begin{algorithmic}[1]
		\Procedure{CW-NGD}{$X \in \mathbb{R}^{P\times n_0}$, $Y\in \mathbb{R}^{P}$}
			\State $S, A\in \left[\mathbb{R}^{P\times n_l} \cup \mathbb{R}^{P\times r_l\times h_l \times c_l}\right]_{l=1,\ldots,L} \gets $ \Call{ForwardProp}{$X$}

			\State {$D_a\in \mathbb{R}^{P\times n_L} \cup \mathbb{R}^{P\times r_L \times h_L \times c_L} \gets$ \Call{Cost$'$}{$A_{[L]}, Y$}}
			\State delete $A_{[L]}$\label{alg:del-1}

			\For{$l$ from $L$ to 2}\label{alg:for-loop-l}
			\State $D_s\in \mathbb{R}^{P\times n_l} \cup \mathbb{R}^{P\times r_l \times h_l \times c_l} \gets D_a \times act_l'(S_{[l]})$\label{alg:state-d-s}
			\State $D_a\in \mathbb{R}^{P\times n_{l-1}} \cup \mathbb{R}^{P\times r_{l-1} \times h_{l-1} \times c_{l-1}} \gets$ \Call{BackProp$_l$}{$D_s$}\label{alg:backprop}
			\If{layer$_l$ is dense}
				\State $D_W\in \mathbb{R}^{P \times n_{l-1} \times n_l} \gets A_{[l-1]}^\intercal \times D_s$
				\State $D_W'\in \mathbb{R}^{n_l \times P \times n_{l-1}} \gets$ \Call{Reshape}{$D_W$}
				\State $F \in \mathbb{R}^{n_l \times n_{l-1} \times n_{l-1}} \gets D_W'^\intercal \times D_W'$
				\State $D_w\in \mathbb{R}^{n_l \times n_{l-1}} \gets$ \Call{Avg$_1$}{$D_W$}\label{alg:state-avg-1}
				\State $D_w'\in \mathbb{R}^{n_l \times 1 \times n_{l-1}} \gets$ \Call{Reshape}{$D_w$}
				\State $U' \in \mathbb{R}^{n_l \times 1 \times n_{l-1}} \gets D_w' \times \left(F + \gamma I\right)^{-1}$\label{alg:gamma-1}
				\State $U \in \mathbb{R}^{n_{l-1} \times n_l} \gets$ \Call{Reshape}{$U'$}
				\State $W_l \gets W_l - \alpha U$
			\ElsIf{layer$_l$ is convolutional}
			\State $D_W\in \mathbb{R}^{P \times k_l \times e_l \times c_{l-1} \times c_l} \gets$ \Call{Conv2DGrad}{$A_{[l-1]}, D_s$}
			\State $D_W'\in \mathbb{R}^{c_l \times P \times k_l e_l c_{l-1}} \gets$ \Call{Reshape}{$D_W$}
			\State $F \in \mathbb{R}^{c_l \times k_l e_l c_{l-1} \times k_l e_l c_{l-1}} \gets D_W'^\intercal \times D_W'$
			\State $D_w\in \mathbb{R}^{k_l \times e_l \times c_{l-1} \times c_l} \gets$ \Call{Avg$_1$}{$D_W$}\label{alg:state-avg-2}
			\State $D_w'\in \mathbb{R}^{c_l \times 1 \times k_l e_l c_{l-1}} \gets$ \Call{Reshape}{$D_w$}
			\State $U' \in \mathbb{R}^{c_l \times 1 \times k_l e_l c_{l-1}} \gets D_w' \times \left(F + \gamma I\right)^{-1}$\label{alg:gamma-2}
			\State $U \in \mathbb{R}^{k_l \times e_l \times c_{l-1} \times c_l} \gets$ \Call{Reshape}{$U'$}
			\State $W_l \gets W_l - \alpha U$
			\EndIf
			\State delete $A_{[l-1]}, S_{[l]}$\label{alg:del-2}
			\EndFor
		\EndProcedure
	\end{algorithmic}
\end{algorithm}
In this subsection, we provide an efficient implementation for CW-NGD.
Next, we evaluate the computational complexity gained by the approximation in CW-NGD.
Finally, we discuss the parallelization capabilities of CW-NGD with SIMD and MIMD.

Pseudocode of CW-NGD is described in Algorithm~\ref{alg:cw-ngd}.
CW-NGD takes two variables, inputs $X$ and labels $Y$, and updates the network weight with our proposed method.
This procedure should be repeated multiple times, where the number of times is determined by the user.
We drop the bias element for simplicity.
In Algorithm~\ref{alg:cw-ngd},
$P, \alpha, \gamma, I$ are batch size, scalar learning rate, scalar damping factor, and unit matrix of appropriate dimension, respectively.
The notation $[]_{l=1,\ldots,L}$ denotes an array of $L$ element numbered by $l$ for example $S$ (or $A$),
and $S_{[l]}$ (or $A_{[l]}$) refers to the $l$-th element of the array $S$ (or $A$).
\Call{ForwardProp}{$X$} is the forward propagation of the network which returns $L$ elements arrays of pre-activation and (post-)activation evaluation of the network.
The network might contain layers of other different types (MaxPooling, Flatten, etc.),
but here we focus on and denote the 2 trainable types of dense ($\mathbb{R}^{P\times n_l}$) and convolutional ($\mathbb{R}^{P \times r_l \times h_l \times c_l}$) layers only.
\Call{Cost$'$}{$A_{[L]}, Y$} is the derivative function of the cost function with respect to the last layer.
The loop from Line~\ref{alg:for-loop-l} iterates from the last layer to the second layer,
and carries out the backpropagation while inverting the FIM to precondition the gradient for updating the weights.
In this for-loop,
$act_l'(S_{[l]})$ returns the Jacobian of the $l$-th layer's activation function,
\Call{BackProp$_l$}{$D_s$} is the backpropagation of the $l$-th layer which returns the derivative with respect to the previous layer (the $(n-1)$-th layer)'s (post-)activation.
For matrix multiplication ($\times$), transpose($^\intercal$), and inversion($^{-1}$) operations, when the operands are tensors of a dimension larger than 2,
the operation is performed on the last two dimensions.
For example, multiplying a $P\times N_1 \times N_2$ tensor with a $P \times N_2 \times N_3$ tensor results a $P \times N_1 \times N_3$ tensor by aggregating results of $P$ multiplications of a $N_1 \times N_2$ matrix with a $N_2 \times N_3$ matrix.
\Call{Avg$_1$}{$A$} is the average of the $A$ tensor along the first dimension.
\Call{Reshape}{$A$} is the reshape function that re-arranges elements of the $A$ tensor into a tensor of a shape specified by the target shape on the left-hand side.

It is worth noting that the backpropagation (Line~\ref{alg:backprop}) calculates the derivative of the cost function with respect to each sample (post-)activation.
This is different from the typical backpropagation in popular libraries that only supports calculating the sum of these derivatives.
Besides, when taking the gradient (Line~\ref{alg:state-avg-1}, Line~\ref{alg:state-avg-2}),
we use the average of the derivative with respect to the weight evaluated at each sample instead of the sum.
The aggregation depends on how the loss across batch samples is aggregated.
If the loss is averaged across samples, then the gradient should be averaged.
Otherwise, when the loss is summed, the gradient should be summed as well.
$\gamma$ is a scalar damping factor that is used (Line~\ref{alg:gamma-1}, Line~\ref{alg:gamma-2}) to prevent the FIM from becoming singular by decreasing its condition number.
This technique is the well-known Tikhonov regularization~\cite{tikhonov} with the Tikhonov matrix being the identity matrix multiplied by $\gamma$.
Tikhonov regularization is proven to be equivalent to imposing a sphere region called the trust region on the update vector~\cite{trust-region}.
So that, sometimes it is also called the trust region method.
Tikhonov regularization is the first-to-use and very simple regularization method used in many quadratic curvature-related problems.
In CW-NGD, we use a relatively small damping factor $\gamma$ and find it highly effective in making the FIMs invertible.

Next, we discuss the computational complexity gained by the approximation in CW-NGD and the memory complexity of the implementation.
The computational complexity of CW-NGD is determined by the most expensive operation of inverting the FIM.
In which, for each layer,
we need to invert $n_l$ matrices of dimension $n_l\times n_l$ (dense) or invert $c_l$ matrices of dimension $k_l e_l c_{l-1}\times k_l e_l c_{l-1}$ (convolutional).
Equivalently, the overall computation complexity is $\mathcal{O}(Ln_l^4) + \mathcal{O}(Lk_l^3e_l^3c_l^4)$,
reduced $\mathcal{O}(max(L^2n_l^2, L^2c_l^2))$ times from $\mathcal{O}(L^3n_l^6) + \mathcal{O}(L^3k_l^3e_l^3c_l^6)$ if we invert the full FIM.

From the memory perspective,
we evaluate the individual FIM of each layer and update the layer's weight during the backpropagation.
We reuse many large-size variables such as $D_W, D_W', F, D_w, D_w', U', U$,
which efficiently reduces memory usage.
Additionally, we remove the pre- and post-activations of each layer immediately after use (Line~\ref{alg:del-1}, Line~\ref{alg:del-2}).

Lastly, we discuss the parallelization capabilities of CW-NGD with SIMD and MIMD.
The SIMD technique can be applied on various matrix operations, especially the FIM inversion (Line~\ref{alg:gamma-1}, Line~\ref{alg:gamma-2}).
Moreover, the inverting operations of the FIM for individual layers and individual components are independent.
So that they can be evaluated in parallel,
and the computational complexity can be further scaled down to $Ln_l$ parallel processes for dense layer (or $Lc_l$ processes for convolutional layer) with each process's computation complexity being $\mathcal{O}(n_l^3)$ ($\mathcal{O}(k_l^3e_l^3c_l^3)$ for convolutional layer).
\begin{figure}
	\center{\includegraphics[width=.45\textwidth]{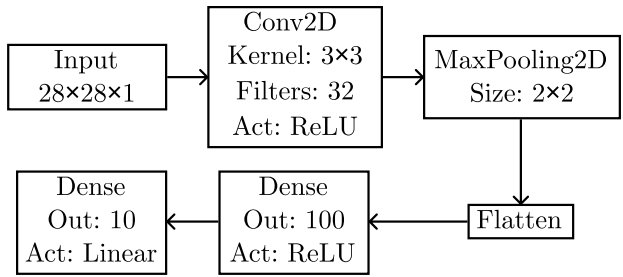}}
	\caption{Model used to train}\label{fig:model}
\end{figure}

\section{Experiment}\label{sec:experiment}
In this section, we describe the details of the conducted experiment including implementation framework, network architecture, dataset, training parameters, and the running environment.

To verify the effectiveness of our proposed method CW-NGD,
we train the model described in Figure~\ref{fig:model} on the MNIST dataset~\cite{mnist} using 3 different training methods:
CW-NGD, KFAC~\cite{kfac-1,kfac-2}, and Adam~\cite{adam}.
The latter 2 methods are the state-of-the-art second-order and first-order optimization methods in the literature, respectively.
The trained model starts with a convolutional layer and a max-pooling layer, followed by 2 dense layers.
Kernel sizes, max pooling window size, and dense layer output sizes are all illustrated in Figure~\ref{fig:model}.
Additionally, we use He Initialization~\cite{he} for all trainable weights.

Our implementation of CW-NGD is based on the latest TensorFlow~\cite{tensorflow} framework version 2.9.1.
For Adam, we use the library call available in the framework.
For KFAC, we use the implementation provided by Tensorflow team at~\cite{kfac-implementation} that requires Tensorflow version 1.15.

The experiment is conducted on the MNIST dataset~\cite{mnist} including 60,000 training samples and 10,000 testing samples,
which are used to determine the training accuracy and validation accuracy described in Section~\ref{sec:results}.
We use the 1024 batch size and train the model in 100 epochs.
In CW-NGD, we set learning rate $\alpha = 1$ and damping factor $\gamma = 0.0001$ according to Algorithm~\ref{alg:cw-ngd}.
For Adam, we use the parameter values and decay strategy provided in~\cite{kfac-implementation} tuned for the CIFAR10 dataset~\cite{cifar10}:
momentum $\beta = 0.97$,
learning rate: initialized at $2.24$ decaying to $0.0001$ at rate $0.9994$,
epsilon: initialized at $0.183$ decaying to $10^{-8}$ at rate $0.9998$.
Similarly, for KFAC, we use the parameter values and decay strategy provided in~\cite{kfac-implementation} tuned for the CIFAR10 dataset~\cite{cifar10}:
momentum $\beta = 0.98$,
learning rate: initialized at $0.23$ decaying to $0.0001$ at rate $0.9989$,
damping factor: initialized at $0.289$ decaying to $10^{-6}$ at rate $0.9997$.
Due to the resource constraint,
we use the hyperparameter values tuned for CIFAR10 to use in this experiment which trains on the MNIST dataset for a heuristic comparison.
This heuristic comparison is sufficient because the hyperparameters only affect the early iterations of the training.

Because the available KFAC implementation is implemented in the legacy Tensorflow version which requires a specific environment (CUDA 10.0, GCC 7.5.0),
to simplify the experiment process,
we run the experiment on 2 machines.
One is equipped with a TITAN RTX GPU, and the other is equipped with a GTX 3080 GPU.
The former is installed with Tensorflow version 2.9.1 (Python 3.10, CUDA 11.2, Cudnn 8.1, GCC 12.1.0) to evaluate CW-NGD and Adam,
while the latter is installed with Tensorflow version 1.15 (Python 3.7, CUDA 10.0, Cudnn 7.4, GCC 7.5.0) to evaluate KFAC.
In 2 machines, Tensorflow automatically selects and makes use of GPGPU for the training.
\pgfplotstableread[col sep=semicolon,trim cells]{
epoch ; cw-ngd ; adam ; kfac
1 ; 0.8797 ; 0.8585 ; 0.9032
2 ; 0.9803 ; 0.9589 ; 0.9568
3 ; 0.9889 ; 0.9739 ; 0.9674
4 ; 0.9931 ; 0.9815 ; 0.9743
5 ; 0.9963 ; 0.986 ; 0.9745
6 ; 0.9979 ; 0.9888 ; 0.9709
7 ; 0.9992 ; 0.9908 ; 0.9779
8 ; 0.9997 ; 0.9923 ; 0.9783
9 ; 0.9999 ; 0.9933 ; 0.9835
10 ; 0.9999 ; 0.9942 ; 0.9872
11 ; 0.9999 ; 0.9949 ; 0.9901
12 ; 0.9999 ; 0.9955 ; 0.9915
13 ; 1 ; 0.9958 ; 0.9919
14 ; 1 ; 0.9955 ; 0.9902
15 ; 1 ; 0.995 ; 0.9903
16 ; 1 ; 0.9959 ; 0.9917
17 ; 1 ; 0.9972 ; 0.9926
18 ; 1 ; 0.9983 ; 0.9939
19 ; 1 ; 0.999 ; 0.9948
20 ; 1 ; 0.9992 ; 0.9959
21 ; 1 ; 0.9994 ; 0.9968
22 ; 1 ; 0.9996 ; 0.9961
23 ; 1 ; 0.9998 ; 0.9964
24 ; 1 ; 0.9999 ; 0.9972
25 ; 1 ; 0.9999 ; 0.9967
26 ; 1 ; 0.9999 ; 0.9973
27 ; 1 ; 0.9999 ; 0.9975
28 ; 1 ; 1 ; 0.998
29 ; 1 ; 1 ; 0.9991
30 ; 1 ; 1 ; 0.9993
31 ; 1 ; 1 ; 0.9996
32 ; 1 ; 1 ; 0.9995
33 ; 1 ; 1 ; 0.9995
34 ; 1 ; 0.9999 ; 0.9996
35 ; 1 ; 1 ; 0.9998
36 ; 1 ; 0.9999 ; 0.9998
37 ; 1 ; 0.9999 ; 0.9999
38 ; 1 ; 0.9999 ; 1
39 ; 1 ; 0.9999 ; 1
40 ; 1 ; 0.9999 ; 1
41 ; 1 ; 0.9999 ; 1
42 ; 1 ; 1 ; 1
43 ; 1 ; 1 ; 1
44 ; 1 ; 1 ; 1
45 ; 1 ; 1 ; 1
46 ; 1 ; 0.9999 ; 1
47 ; 1 ; 1 ; 1
48 ; 1 ; 0.9999 ; 1
49 ; 1 ; 1 ; 1
50 ; 1 ; 1 ; 1
51 ; 1 ; 1 ; 1
52 ; 1 ; 1 ; 1
53 ; 1 ; 1 ; 1
54 ; 1 ; 1 ; 1
55 ; 1 ; 1 ; 1
56 ; 1 ; 1 ; 1
57 ; 1 ; 1 ; 1
58 ; 1 ; 1 ; 1
59 ; 1 ; 1 ; 1
60 ; 1 ; 1 ; 1
61 ; 1 ; 1 ; 1
62 ; 1 ; 1 ; 1
63 ; 1 ; 1 ; 1
64 ; 1 ; 1 ; 1
65 ; 1 ; 1 ; 1
66 ; 1 ; 1 ; 1
67 ; 1 ; 1 ; 1
68 ; 1 ; 1 ; 1
69 ; 1 ; 1 ; 1
70 ; 1 ; 1 ; 1
71 ; 1 ; 1 ; 1
72 ; 1 ; 1 ; 1
73 ; 1 ; 1 ; 1
74 ; 1 ; 1 ; 1
75 ; 1 ; 1 ; 1
76 ; 1 ; 1 ; 1
77 ; 1 ; 1 ; 1
78 ; 1 ; 1 ; 1
79 ; 1 ; 1 ; 1
80 ; 1 ; 1 ; 1
81 ; 1 ; 1 ; 1
82 ; 1 ; 1 ; 1
83 ; 1 ; 1 ; 1
84 ; 1 ; 1 ; 1
85 ; 1 ; 1 ; 1
86 ; 1 ; 1 ; 1
87 ; 1 ; 1 ; 1
88 ; 1 ; 1 ; 1
89 ; 1 ; 1 ; 1
90 ; 1 ; 1 ; 1
91 ; 1 ; 1 ; 1
92 ; 1 ; 1 ; 1
93 ; 1 ; 1 ; 1
94 ; 1 ; 1 ; 1
95 ; 1 ; 1 ; 1
96 ; 1 ; 1 ; 1
97 ; 1 ; 1 ; 1
98 ; 1 ; 1 ; 1
99 ; 1 ; 1 ; 1
100 ; 1 ; 1 ; 1
}\TrainingAccuracy

\pgfplotstableread[col sep=semicolon,trim cells]{
epoch ; cw-ngd ; adam ; kfac
1 ; 0.9538 ; 0.9464 ; 0.9463
2 ; 0.9643 ; 0.9683 ; 0.9491
3 ; 0.9809 ; 0.9747 ; 0.9677
4 ; 0.983 ; 0.98 ; 0.9618
5 ; 0.9803 ; 0.9818 ; 0.9615
6 ; 0.987 ; 0.9821 ; 0.9696
7 ; 0.9851 ; 0.9829 ; 0.9583
8 ; 0.9875 ; 0.9836 ; 0.9736
9 ; 0.9893 ; 0.9836 ; 0.9734
10 ; 0.9881 ; 0.9838 ; 0.9701
11 ; 0.9884 ; 0.9831 ; 0.9764
12 ; 0.9889 ; 0.9831 ; 0.9731
13 ; 0.9884 ; 0.9825 ; 0.9754
14 ; 0.9883 ; 0.9818 ; 0.9771
15 ; 0.9886 ; 0.9828 ; 0.9743
16 ; 0.9886 ; 0.9836 ; 0.9767
17 ; 0.9889 ; 0.9844 ; 0.9811
18 ; 0.9892 ; 0.9861 ; 0.9819
19 ; 0.9896 ; 0.9861 ; 0.9793
20 ; 0.9895 ; 0.9855 ; 0.9814
21 ; 0.9895 ; 0.9847 ; 0.9785
22 ; 0.9893 ; 0.9851 ; 0.9794
23 ; 0.9894 ; 0.9854 ; 0.9816
24 ; 0.9896 ; 0.9853 ; 0.981
25 ; 0.9897 ; 0.9853 ; 0.9784
26 ; 0.9895 ; 0.9853 ; 0.9798
27 ; 0.9896 ; 0.9855 ; 0.9785
28 ; 0.9896 ; 0.9852 ; 0.9819
29 ; 0.9894 ; 0.985 ; 0.9798
30 ; 0.9898 ; 0.9849 ; 0.9823
31 ; 0.9897 ; 0.9856 ; 0.9823
32 ; 0.9899 ; 0.9853 ; 0.9829
33 ; 0.9898 ; 0.9849 ; 0.9817
34 ; 0.9897 ; 0.9851 ; 0.9818
35 ; 0.9897 ; 0.9858 ; 0.982
36 ; 0.9896 ; 0.9863 ; 0.9826
37 ; 0.9897 ; 0.9847 ; 0.9825
38 ; 0.9898 ; 0.9859 ; 0.983
39 ; 0.9898 ; 0.9857 ; 0.9823
40 ; 0.9898 ; 0.9856 ; 0.9832
41 ; 0.9897 ; 0.9857 ; 0.9832
42 ; 0.9895 ; 0.9865 ; 0.9826
43 ; 0.9899 ; 0.9862 ; 0.9827
44 ; 0.9897 ; 0.9867 ; 0.9834
45 ; 0.9898 ; 0.9864 ; 0.9837
46 ; 0.9897 ; 0.9865 ; 0.9835
47 ; 0.9896 ; 0.9872 ; 0.9833
48 ; 0.9899 ; 0.9866 ; 0.9833
49 ; 0.9898 ; 0.9864 ; 0.9832
50 ; 0.9899 ; 0.9855 ; 0.9832
51 ; 0.9898 ; 0.9867 ; 0.9832
52 ; 0.9898 ; 0.9869 ; 0.9832
53 ; 0.9898 ; 0.9872 ; 0.9832
54 ; 0.9898 ; 0.9874 ; 0.9833
55 ; 0.9899 ; 0.9873 ; 0.9833
56 ; 0.9898 ; 0.9873 ; 0.9833
57 ; 0.9897 ; 0.9873 ; 0.9833
58 ; 0.9898 ; 0.9873 ; 0.9833
59 ; 0.9899 ; 0.9873 ; 0.9832
60 ; 0.9898 ; 0.9871 ; 0.9832
61 ; 0.9897 ; 0.9871 ; 0.9832
62 ; 0.9898 ; 0.9871 ; 0.9832
63 ; 0.9898 ; 0.9871 ; 0.9832
64 ; 0.9898 ; 0.9871 ; 0.9832
65 ; 0.9897 ; 0.9871 ; 0.9832
66 ; 0.9897 ; 0.9871 ; 0.9832
67 ; 0.9898 ; 0.9871 ; 0.9832
68 ; 0.9896 ; 0.987 ; 0.9832
69 ; 0.9896 ; 0.987 ; 0.9832
70 ; 0.9896 ; 0.987 ; 0.9832
71 ; 0.9895 ; 0.9869 ; 0.9832
72 ; 0.9896 ; 0.9869 ; 0.9832
73 ; 0.9895 ; 0.9869 ; 0.9832
74 ; 0.9895 ; 0.9868 ; 0.9832
75 ; 0.9895 ; 0.9868 ; 0.9832
76 ; 0.9895 ; 0.9868 ; 0.9832
77 ; 0.9895 ; 0.9868 ; 0.9832
78 ; 0.9897 ; 0.9868 ; 0.9832
79 ; 0.9894 ; 0.9868 ; 0.9832
80 ; 0.9896 ; 0.9867 ; 0.9832
81 ; 0.9896 ; 0.9867 ; 0.9832
82 ; 0.9896 ; 0.9867 ; 0.9832
83 ; 0.9895 ; 0.9867 ; 0.9832
84 ; 0.9895 ; 0.9868 ; 0.9832
85 ; 0.9894 ; 0.9867 ; 0.9832
86 ; 0.9894 ; 0.9867 ; 0.9832
87 ; 0.9895 ; 0.9867 ; 0.9832
88 ; 0.9893 ; 0.9867 ; 0.9832
89 ; 0.9894 ; 0.9868 ; 0.9832
90 ; 0.9894 ; 0.9867 ; 0.9832
91 ; 0.9894 ; 0.9866 ; 0.9832
92 ; 0.9895 ; 0.9868 ; 0.9832
93 ; 0.9895 ; 0.9867 ; 0.9832
94 ; 0.9894 ; 0.9867 ; 0.9832
95 ; 0.9894 ; 0.9868 ; 0.9832
96 ; 0.9893 ; 0.9868 ; 0.9832
97 ; 0.9894 ; 0.9868 ; 0.9832
98 ; 0.9894 ; 0.987 ; 0.9832
99 ; 0.9893 ; 0.9869 ; 0.9832
100 ; 0.9896 ; 0.9868 ; 0.9832
}\ValidationAccuracy
\section{Results}\label{sec:results}
\begin{table}[h!]
	\centering
	\begin{tabular}{l r r r}
		\toprule
		Optimization & CW-NGD & Adam & KFAC \\
		\midrule
		\midrule
		\Centerstack[l]{1st training iteration\\training accuracy} & 87.97\% & 85.85\% & 90.32\% \\
		\midrule
		\Centerstack[l]{2nd training iteration\\ training accuracy} & 98.03\% & 95.89\% & 95.68\% \\
		\midrule
		\Centerstack[l]{Number of iterations\\needed for convergence} & 13 & 49 & 38 \\
		\midrule
		training accuracy & 100\% & 100\% & 100\% \\
		\midrule
		validation accuracy & 98.96\% & 98.68\% & 98.32\% \\
		\bottomrule
	\end{tabular}
	\caption{Comparison among 3 optimization methods}\label{tab:optimizations}
\end{table}

\begin{figure}
	\begin{subfigure}{4.5cm}
		\centering
		\begin{tikzpicture}
			\begin{axis}[
				height=5cm,width=6cm,
				xlabel={Iteration}, xlabel near ticks,
				ylabel={Training Accuracy}, ylabel near ticks,
				legend style={at={(1.3, 1.)},anchor=north},
				legend entries={{CW-NGD}, {Adam}, {KFAC}},
				mark repeat=10,
				xtick={0, 20, 40, 60, 80, 100},
				mark options={solid,scale=1},
			]
				\addplot[mark=square] table [x=epoch, y={cw-ngd}] {\TrainingAccuracy};
				\addplot[mark=o] table [x=epoch, y={adam}] {\TrainingAccuracy};
				\addplot[mark=triangle] table [x=epoch, y={kfac}] {\TrainingAccuracy};
			\end{axis}
		\end{tikzpicture}
		\caption{\scriptsize Training Accuracy}\label{fig:training-accuracy}
	\end{subfigure}
	\par\bigskip
	\begin{subfigure}{4.5cm}
		\centering
		\begin{tikzpicture}
			\begin{axis}[
				height=5cm,width=6cm,
				xlabel={Iteration}, xlabel near ticks,
				ylabel={Validation Accuracy}, ylabel near ticks,
				legend style={at={(1.3, 1.)},anchor=north},
				legend entries={{CW-NGD}, {Adam}, {KFAC}},
				xtick={0, 20, 40, 60, 80, 100},
				mark repeat=50,
				mark options={solid,scale=1},
			]
				\addplot[mark=square] table [x=epoch, y={cw-ngd}] {\ValidationAccuracy};
				\addplot[mark=o] table [x=epoch, y={adam}] {\ValidationAccuracy};
				\addplot[mark=triangle] table [x=epoch, y={kfac}] {\ValidationAccuracy};
			\end{axis}
		\end{tikzpicture}
		\caption{\scriptsize Validation Accuracy}\label{fig:validation-accuracy}
	\end{subfigure}
	\caption{Accuracy comparison among 3 optimization methods (higher is better)}\label{fig:accuracy}
\end{figure}
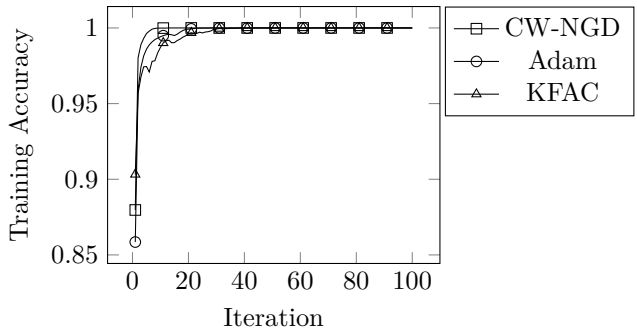
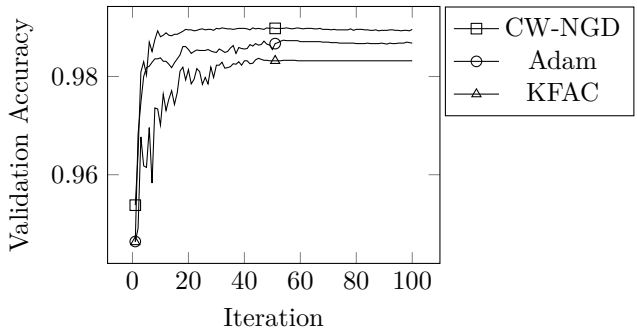

In this section, we discuss the experiment results of 3 optimization methods (CW-NGD, Adam, KFAC) based on the graphs shown in Figure~\ref{fig:accuracy},
with several statistic results in Table~\ref{tab:optimizations}.

Figure~\ref{fig:training-accuracy} gives an overview of the training accuracy of the three optimization methods.
The higher curve of CW-NGD indicates that it performs better than the other two methods by converging within fewer iterations.
Specifically, CW-NGD converges at the 13th iteration, while the other two methods converge at the 49th and 38th iterations (Table~\ref{tab:optimizations}).
In other words, CW-NGD requires nearly one-third fewer iterations to converge than the other two methods.

Additionally, in Figure~\ref{fig:training-accuracy}, CW-NGD curve smoothly and monotonically rises without fluctuating.
This result can be partially explained by the absence of decaying strategy in CW-NGD training.
In contrast, the other two methods' graphs fluctuate due to the involvement of decaying strategies.
CW-NGD achieves a higher training accuracy than the other two methods even before the first fluctuation,
implying that CW-NGD produces a higher precondition for the gradient update compared to the other two methods.
Table~\ref{tab:optimizations} shows that CW-NGD starts with a lower training accuracy than KFAC, better than Adam,
then immediately gets a better training accuracy than both from the second iteration.

Regarding the validation accuracy, Figure~\ref{fig:validation-accuracy} clearly indicates that CW-NGD outperforms the other two methods.
To be precise, CW-NGD's final validation accuracy is 98.96\% while the other two methods' final validation accuracies are lower and are 98.68\% and 98.32\%, respectively (Table~\ref{tab:optimizations}).
Moreover, CW-NGD's validation accuracy curve fluctuates less than the other two methods suggesting that CW-NGD's produced gradient update precondition is more stable than the other two methods.
\section{Conclusion}\label{sec:conclusion}
In this work, we propose a novel method, named Component-Wise Natural Gradient Descent (CW-NGD), for training neural networks.
CW-NGD is a second-order optimization that works by approximating the Fisher Information Matrix (FIM) required during the training as a block-diagonal matrix.
Each block matrix in the diagonal corresponds to the FIM of a group of network weights in a layer.
We introduce grouping strategies for CW-NGD that efficiently works on dense and convolutional layers.
Our investigations show that CW-NGD runs in a reasonable time while preserving the accuracy of the model at a high level.

We provide the detail of an efficient implementation of CW-NGD that generates the FIM, inverts it, and updates the weights while
doing the backpropagation.
The implementation also reduces memory usage by reusing several large-size variables.

In an experiment that compares the number of iterations required for training convergence,
we show that CW-NGD outperforms KFAC and Adam, 2 state-of-the-art second-order and first-order optimization methods in the literature, respectively.
Specifically, CW-NGD only requires one-third of the iterations required by KFAC or Adam to converge.

In another experiment,
we compare the validation accuracy of the 3 methods.
The results reveal that CW-NGD clearly has the highest validation accuracy.

From the performance perspective,
theoretically, the matrix inversion of the FIM can be divided into independent subproblems corresponding to the individual FIM of a component,
each of which can be solved by a single process.
This indicates the potential for a high parallelization level of CW-NGD.
In the future,
we will investigate a distributed solution of CW-NGD.
Additionally, we will also extend CW-NGD to support layers other than dense and convolution.
Besides, we are going to evaluate CW-NGD on more sophisticated datasets such as CIFAR10~\cite{cifar10}, CIFAR100~\cite{cifar10}, ImageNet~\cite{imagenet}, etc.

	\bibliographystyle{IEEEtran}
	\bibliography{ref}

\end{document}